\documentclass[10pt,twocolumn,letterpaper]{article}

\usepackage{cvpr}
\usepackage{times}
\usepackage{epsfig}
\usepackage{graphicx}
\usepackage{amsmath}
\usepackage{amssymb}
\usepackage{enumitem} 

\DeclareMathOperator*{\argmax}{argmax}
\usepackage{multirow}
\usepackage{booktabs}
\usepackage{subfigure}
\usepackage{caption}
\usepackage{pifont}
\usepackage{soul}

\usepackage[pagebackref=true,breaklinks=true,letterpaper=true,colorlinks,bookmarks=false]{hyperref}

\cvprfinalcopy 

\def\cvprPaperID{10042} 

\newcommand{\rb}{\rotatebox{90}}%
\newcommand{\cmark}{\ding{51}}%
\newcommand{\xmark}{\ding{55}}%

\ifcvprfinal\fi
\begin{document}

\title{Learning Memory-guided Normality for Anomaly Detection}

\author{Hyunjong Park\thanks{Equal contribution.~$^\dagger$Corresponding author.} \quad\quad\quad Jongyoun Noh\footnotemark[1] \quad\quad\quad Bumsub Ham\textsuperscript{$\dagger$}\vspace*{0.1cm}\\
School of Electrical and Electronic Engineering, Yonsei University\\
}

\maketitle
\thispagestyle{empty}

\begin{abstract}
\vspace{-0.3cm}
We address the problem of anomaly detection, that is, detecting anomalous events in a video sequence. Anomaly detection methods based on convolutional neural networks~(CNNs) typically leverage proxy tasks, such as reconstructing input video frames, to learn models describing normality without seeing anomalous samples at training time, and quantify the extent of abnormalities using the reconstruction error at test time. The main drawbacks of these approaches are that they do not consider the diversity of normal patterns explicitly, and the powerful representation capacity of CNNs allows to reconstruct abnormal video frames. To address this problem, we present an unsupervised learning approach to anomaly detection that considers the diversity of normal patterns explicitly, while lessening the representation capacity of CNNs. To this end, we propose to use a memory module with a new update scheme where items in the memory record prototypical patterns of normal data. We also present novel feature compactness and separateness losses to train the memory, boosting the discriminative power of both memory items and deeply learned features from normal data. Experimental results on standard benchmarks demonstrate the effectiveness and efficiency of our approach, which outperforms the state of the art.

\end{abstract}

 	\begin{figure}[t]
		\centering
	\captionsetup{font={small}}
		\includegraphics[width=0.9\linewidth]{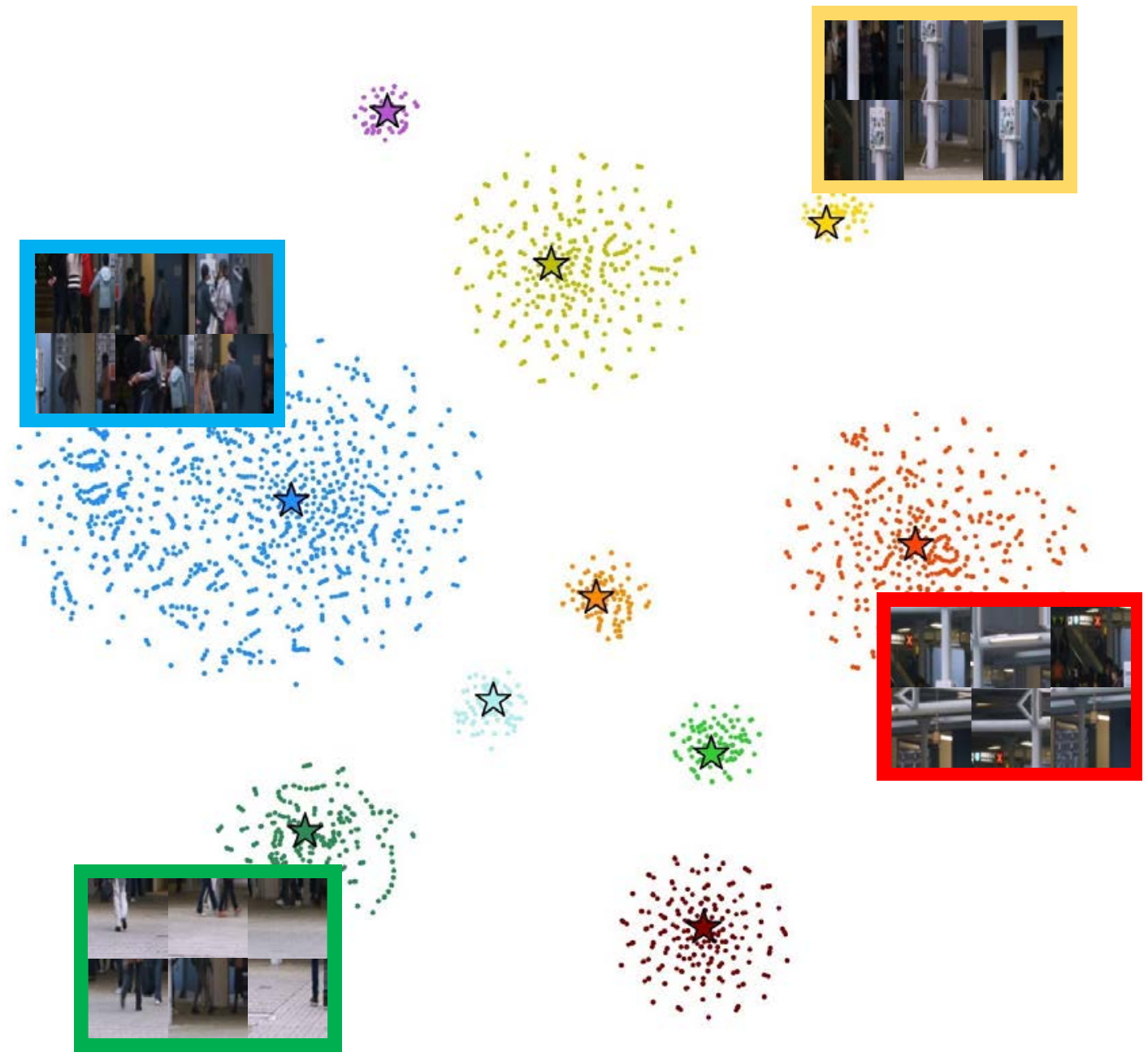}
\vspace{-0.1cm}
		\caption{Distributions of features and memory items of our model on CUHK Avenue~\cite{lu2013abnormal}. The features and items are shown in points and stars, respectively. The points with the same color are mapped to the same item. The items in the memory capture diverse and prototypical patterns of normal data. The features are highly discriminative and similar image patches are clustered well. (Best viewed in color.)}
		\vspace{-0.3cm}
		\label{fig:teaser}
    \end{figure}

\vspace{-0.2cm}
\section{Introduction}
\vspace{-0.1cm}
The problem of detecting abnormal events in video sequences,~\eg,~vehicles on sidewalks, has attracted significant attention over the last decade, which is particularly important for surveillance and fault detection systems. It is extremely challenging for a number of reasons: First, anomalous events are determined differently according to circumstances. Namely, the same activity could be normal or abnormal~(\eg,~holding a knife in the kitchen or in the park). Manually annotating anomalous events is in this context labor intensive. Second, collecting anomalous datasets requires a lot of effort, as anomalous events rarely happen in real-life situations. Anomaly detection is thus typically deemed to be an unsupervised learning problem, aiming at learning a model describing normality without anomalous samples. At test time, events and activities not described by the model are then considered as anomalies. 

There are many attempts to model normality in video sequences using unsupervised learning approaches. At training time, given normal video frames as inputs, they typically extract feature representations and try to reconstruct the inputs again. The video frames of large reconstruction errors are then treated as anomalies at test time. This assumes that abnormal samples are not reconstructed well, as the models have never seen them during training. Recent methods based on convolutional neural networks~(CNNs) exploit an autoencoder~(AE)~\cite{bengio2007greedy,kingma2013auto}. The powerful representation capacity of CNNs allows to extract better feature representations. The CNN features from abnormal frames, on the other hand, are likely to be reconstructed by combining those of normal ones~\cite{liu2018future,gong2019memorizing}. In this case, abnormal frames have low reconstruction errors, often occurring when a majority of the abnormal frames are normal~(\eg,~pedestrians in a park). In order to lessen the capacity of CNNs, a video prediction framework~\cite{liu2018future} is introduced that minimizes the difference between a predicted future frame and its ground truth. The drawback of these methods~\cite{bengio2007greedy,kingma2013auto,liu2018future} is that they do not detect anomalies directly~\cite{ruff2018deep}. They instead leverage proxy tasks for anomaly detection,~\eg,~reconstructing input frames~\cite{bengio2007greedy,kingma2013auto} or predicting future frames~\cite{liu2018future}, to extract general feature representations rather than normal patterns. To overcome this problem, Deep SVDD~\cite{ruff2018deep} exploits the one-class classification objective to map normal data into a hypersphere. Specifically, it minimizes the volume of the hypersphere such that normal samples are mapped closely to the center of the sphere. Although a single center of the sphere represents a universal characteristic of normal data, this does not consider various patterns of normal samples.


We present in this paper an unsupervised learning approach to anomaly detection in video sequences considering the diversity of normal patterns. We assume that a single prototypical feature is not enough to represent various patterns of normal data. That is, multiple prototypes~(\ie,~modes or centroids of features) exist in the feature space of normal video frames~(Fig.~\ref{fig:teaser}). To implement this idea, we propose a memory module for anomaly detection, where individual items in the memory correspond to prototypical features of normal patterns. We represent video frames using the prototypical features in the memory items, lessening the capacity of CNNs. To reduce the intra-class variations of CNN features, we propose a feature compactness loss, mapping the features of a normal video frame to the nearest item in the memory and encouraging them to be close. Simply updating memory items and extracting CNN features alternatively in turn give a degenerate solution, where all items are similar and thus all features are mapped closely in the embedding space. To address this problem, we propose a feature separateness loss. It minimizes the distance between each feature and its nearest item, while maximizing the discrepancy between the feature and the second nearest one, separating individual items in the memory and enhancing the discriminative power of the features and memory items. We also introduce an update strategy to prevent the memory from recording features of anomalous samples at test time. To this end, we propose a weighted regular score measuring how many anomalies exist within a video frame, such that the items are updated only when the frame is determined as a normal one. Experimental results on standard benchmarks, including UCSD Ped2~\cite{li2013anomaly}, CUHK Avenue~\cite{lu2013abnormal} and ShanghaiTech~\cite{luo2017revisit}, demonstrate the effectiveness and efficiency of our approach, outperforming the state of the art. 

The main contributions of this paper can be summarized as follows: \vspace{-0.2cm}
	\begin{itemize}[leftmargin=*]
		\item We propose to use multiple prototypes to represent the diverse patterns of normal video frames for unsupervised anomaly detection. To this end, we introduce a memory module recording prototypical patterns of normal data on the items in the memory.
\vspace{-0.2cm}
		\item We propose feature compactness and separateness losses to train the memory, ensuring the diversity and discriminative power of the memory items. We also present a new update scheme of the memory, when both normal and abnormal samples exist at test time. 
\vspace{-0.2cm}
		\item We achieve a new state of the art on standard benchmarks for unsupervised anomaly detection in video sequences. We also provide an extensive experimental analysis with ablation studies. 
	\end{itemize}
\vspace{-0.2cm}
Our code and models are available online:~\url{https://cvlab.yonsei.ac.kr/projects/MNAD}.

\vspace{-0.1cm}
\section{Related work}
\vspace{-0.1cm}
	\paragraph{Anomaly detection.}
		Many works formulate anomaly detection as an unsupervised learning problem, where anomalous data are not available at training time. They typically adopt reconstructive or discriminative approaches to learn models describing normality. Reconstructive models encode normal patterns using representation learning methods such as an AE~\cite{zhai2016deep,sabokrou2015real}, a sparse dictionary learning~\cite{cong2011sparse,zhao2011online,lu2013abnormal}, and a generative model~\cite{vaswani2005shape}. Discriminative models characterize the statistical distributions of normal samples and obtain decision boundaries around the normal instances~\eg,~using Markov random field~(MRF)~\cite{kim2009observe}, a mixture of dynamic textures~(MDT)~\cite{mahadevan2010anomaly}, Gaussian regression~\cite{cheng2015video}, and one-class classification~\cite{scholkopf2001estimating,ma2003time,kaltsa2015swarm}. These approaches, however, often fail to capture the complex distributions of high-dimensional data such as images and videos~\cite{chalapathy2019deep}.

	\begin{figure}[t]
		\centering
	\captionsetup{font={small}}
		\includegraphics[width=0.85\linewidth]{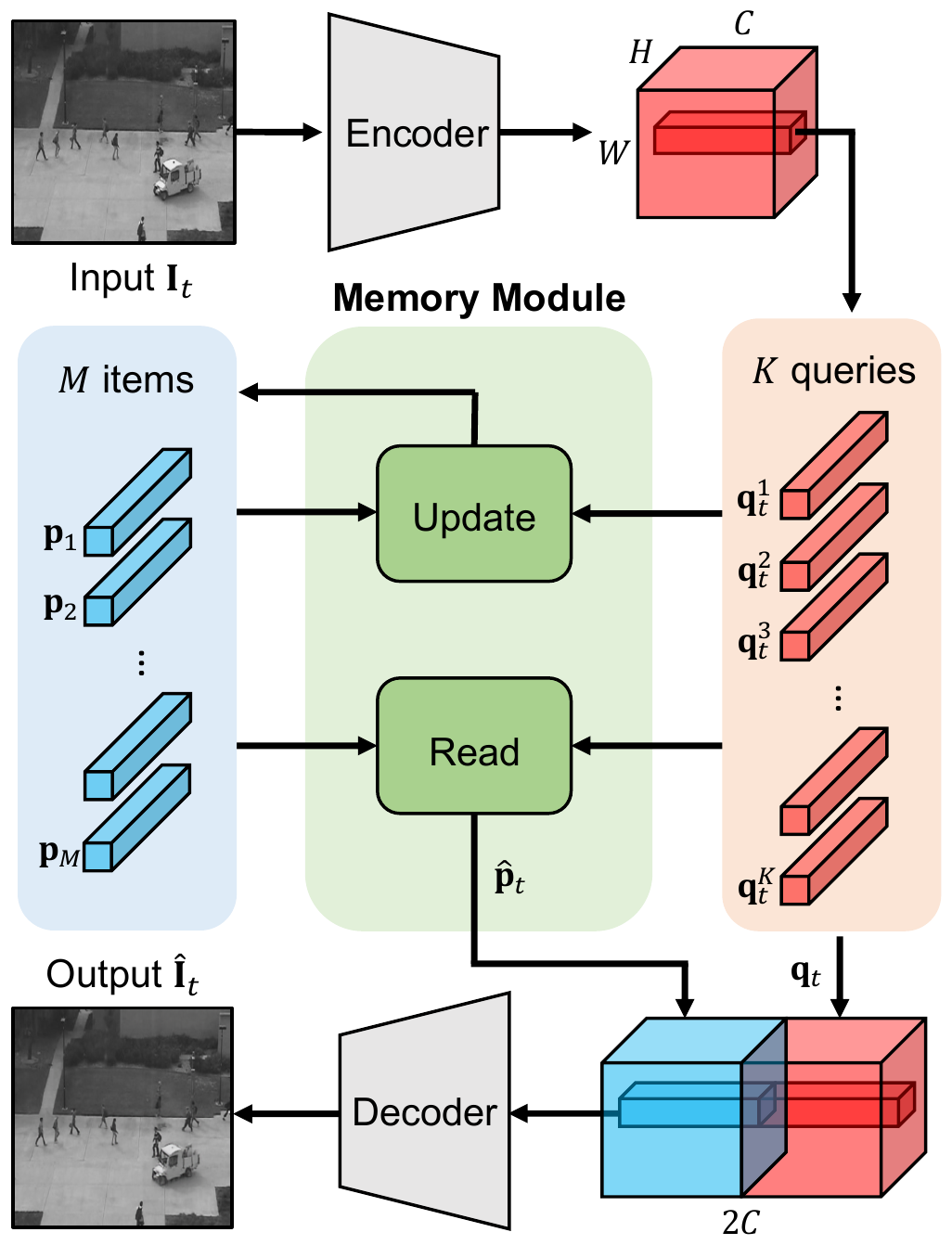}
\vspace{-0.2cm}
		\caption{Overview of our framework for reconstructing a video frame. Our model mainly consists of three parts: an encoder, a memory module, and a decoder. The encoder extracts a query map~$\mathbf{q}_t$ of size~$H\times W \times C$ from an input video frame~${\bf{I}}_t$ at time~$t$. The memory module performs reading and updating items~$\mathbf{p}_m$ of size~$1\times 1\times C$ using queries~$\mathbf{q}_t^k$ of size~$1\times 1\times C$, where the numbers of items and queries are $M$ and $K$, respectively, and $K=H\times W$. The query map~$\mathbf{q}_t$ is concatenated with the aggregated~(\ie,~read) items~$\hat {\bf{p}}_t$. The decoder then inputs them to reconstruct the video frame~$\hat {\bf{I}}_t$. For the prediction task, we input four successive video frames to predict the fifth one. (Best viewed in color.)}
		\label{fig:overview}
\vspace{-0.3cm}
    \end{figure}

		CNNs have allowed remarkable advances in anomaly detection over the last decade. Many anomaly detection methods leverage reconstructive models~\cite{hasan2016learning,luo2017revisit,chong2017abnormal,ravanbakhsh2017abnormal} exploiting feature representations from~\eg,~a convolutional AE~(Conv-AE)~\cite{hasan2016learning}, a 3D Conv-AE~\cite{zhao2017spatio}, a recurrent neural network~(RNN)~\cite{medel2016anomaly,luo2017revisit,luo2017remembering}, and a generative adversarial network~(GAN)~\cite{ravanbakhsh2017abnormal}. Although CNN-based methods outperform classical approaches by large margins, they even reconstruct anomalous samples with a combination of normal ones, mainly due to the representation capacity of CNNs. This problem can be alleviated by using predictive or discriminative models~\cite{liu2018future,ruff2018deep}. The work of~\cite{liu2018future} assumes that anomalous frames in video sequences are unpredictable, and trains a network for predicting future frames rather than the input itself~\cite{liu2018future}. It achieves a remarkable performance gain over reconstructive models, but at the cost of runtime for estimating optical flow between video frames. It also requires ground-truth optical flow to train a sub-network for computing flow fields. Deep SVDD~\cite{ruff2018deep} leverages CNNs as mapping functions that transform normal data into the center of the hypersphere, whereas forcing anomalous samples to fall outside the sphere, using the one-class classification objective. Our method also lessens the representation capacity of CNNs but using a different way. We reconstruct or predict a video frame with a combination of items in the memory, rather than using CNN features directly from an encoder, while considering various patterns of normal data. In case of future frame prediction, our model does not require computing optical flow, and thus it is much faster than the current method~\cite{liu2018future}. Deep-Cascade~\cite{sabokrou2017deep} detects various normal patches using cascaded deep networks. In contrast, our method leverages memory items to record the normal pattern explicitly even in test sequences. Concurrent to our method, Gong~\emph{et al}.~introduce a memory-augmented autoencoder~(MemAE) for anomaly detection~\cite{gong2019memorizing}. It also uses CNN features but using a 3D Conv-AE to retrieve relevant memory items that record normal patterns, where the items are updated during training only. Unlike this approach, our model better records \emph{diverse and discriminative} normal patterns by separating memory items explicitly using feature compactness and separateness losses, enabling using a small number of items compared to MemAE~(10 vs 2,000 for MemAE). We also update the memory at test time, while discriminating anomalies simultaneously, suggesting that our model also memorizes normal patterns of test data.

\vspace{-0.5cm}
	\paragraph{Memory networks.}
	There are a number of attempts to capture long-term dependencies in sequential data. Long short-term memory~(LSTM)~\cite{hochreiter1997long} addresses this problem using local memory cells, where hidden states of the network record information in the past partially. The memorization performance is, however, limited, as the size of the cell is typically small and the knowledge in the hidden state is compressed. To overcome the limitation, memory networks~\cite{weston2015memory} have recently been introduced. It uses a global memory that can be read and written to, and performs a memorization task better than classical approaches. The memory networks, however, require layer-wise supervision to learn models, making it hard to train them using standard backpropagation. More recent works use continuous memory representations~\cite{sukhbaatar2015end} or key-value pairs~\cite{miller2016key} to read/write memories, allowing to train the memory networks end-to-end. Several works adopt the memory networks for computer vision tasks including visual question answering~\cite{kumar2016ask,fan2019heterogeneous}, one-shot learning~\cite{santoro2016meta,kaiser2017learning,cai2018memory}, image generation~\cite{zhu2019dm}, and video summarization~\cite{lee2018memory}. Our work also exploits a memory module but for anomaly detection with a different memory updating strategy. We record various patterns of normal data to individual items in the memory, and consider each item as a prototypical feature.

\vspace{-0.2cm}    
\section{Approach}
\vspace{-0.1cm}

		We show in Fig.~\ref{fig:overview} an overview of our framework. We reconstruct input frames or predict future ones for unsupervised anomaly detection. Following~\cite{liu2018future}, we input four successive video frames to predict the fifth one for the prediction task. As the prediction can be considered as a reconstruction of the future frame using previous ones, we use almost the same network architecture with the same losses for both tasks. We describe hereafter our approach for the reconstruction task in detail. 

Our model mainly consists of three components: an encoder, a memory module, and a decoder. The encoder inputs a normal video frame and extracts query features. The features are then used to retrieve prototypical normal patterns in the memory items and to update the memory. We feed the query features and memory items aggregated~(\ie,~read) to the decoder for reconstructing the input video frame. We train our model using reconstruction, feature compactness, and feature separateness losses end-to-end. At test time, we use a weighted regular score in order to prevent the memory from being updated by abnormal video frames. We compute the discrepancies between the input frame and its reconstruction and the distances between the query feature and the nearest item in the memory to quantify the extent of abnormalities in a video frame.

 	\begin{figure}[t]
	\captionsetup{font={small}}
		\centering
			\includegraphics[width=0.85\linewidth]{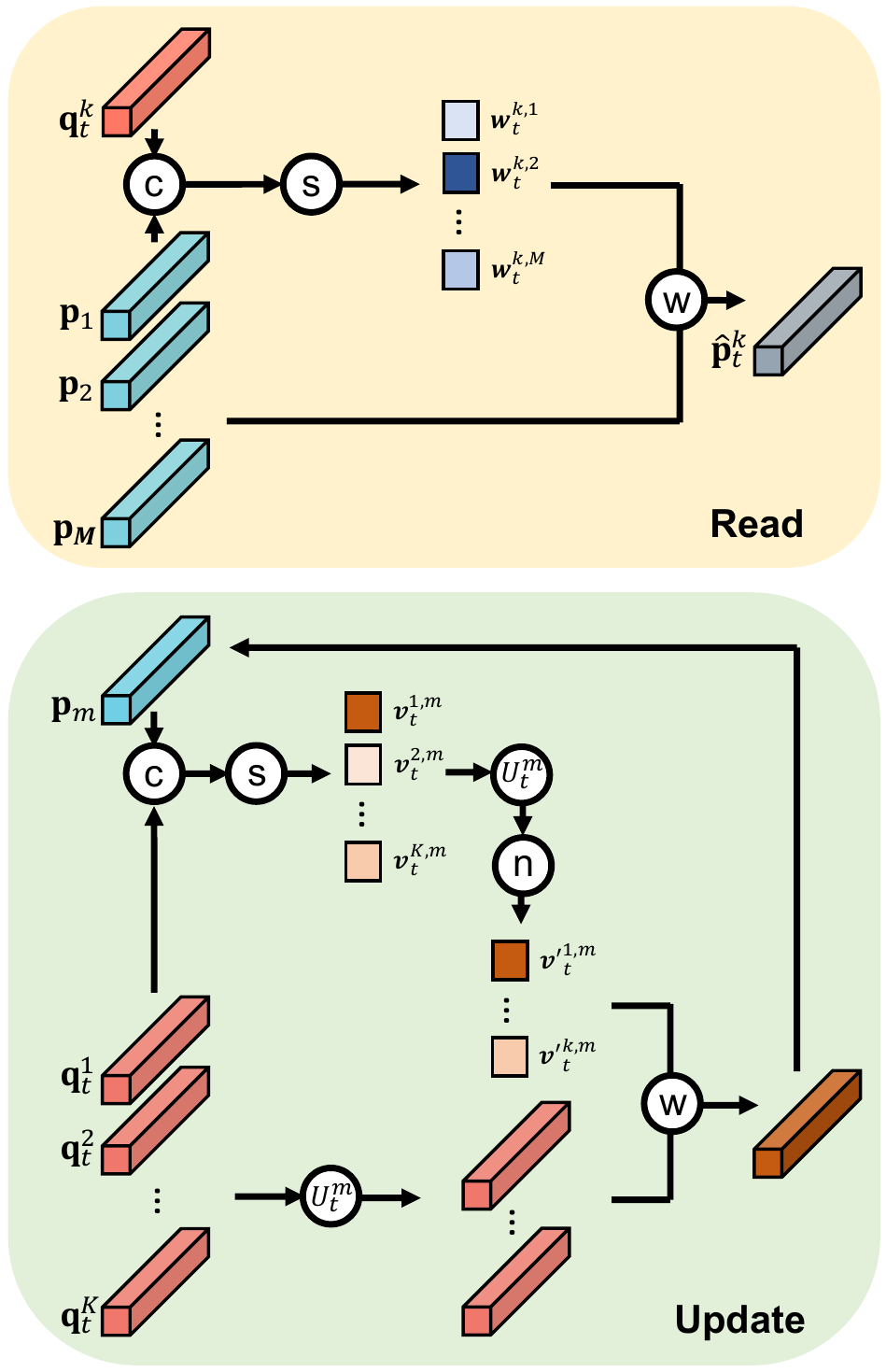}
\vspace{-0.2cm}
		\caption{Illustration of reading and updating the memory. To read items in the memory, we compute matching probabilities~$w_t^{k,m}$ in~\eqref{eq:w} between the query~$\mathbf{q}_t^k$ and items~($\mathbf{p}_1, \dots \mathbf{p}_M$), and apply a weighted average of the items with the probabilities to obtain the feature~$\hat{\mathbf{p}}_t^k$. To update the items, we compute another matching probabilities~$v_t^{k,m}$ in~\eqref{eq:v} between the item~$\mathbf{p}_m$ and the queries~($\mathbf{q}_t^1, \dots \mathbf{q}_t^K$). We then compute a weighted average of the queries in the set~$U_t^m$ with the corresponding probabilities, and add it to the initial item~$\mathbf{p}_m$ in~\eqref{eq:update}. c: cosine similarities; s: a softmax function; w: a weighted average; n: max normalization; $U_t^m$: a set of indices for the $m$-th memory item. See text for details. (Best viewed in color.)}
		\vspace{-0.3cm}
		\label{fig:memory}
    \end{figure}

    \vspace{-0.1cm}
	\subsection{Network architecture}\label{sec:network}
\vspace{-0.1cm}

		\subsubsection{Encoder and decoder}\vspace{-0.2cm}
			We exploit the U-Net architecture~\cite{ronneberger2015u}, widely used for the tasks of reconstruction and future frame prediction~\cite{liu2018future}, to extract feature representations from input video frames and to reconstruct the frames from the features. Differently, we remove the last batch normalization~\cite{ioffe2015batch} and ReLU layers~\cite{krizhevsky2012imagenet} in the encoder, as the ReLU cuts off negative values, restricting diverse feature representations. We instead add an L2 normalization layer to make the features have a common scale. Skip connections in the U-Net architecture may not be able to extract useful features from the video frames especially for the reconstruction task, and our model may learn to copy the inputs for the reconstruction. We thus remove the skip connections for the reconstruction task, while retaining them for predicting future frames. We denote by~${\bf{I}}_t$ and~${\bf{q}}_t$ a video frame and a corresponding feature~(\ie,~a query) from the encoder at time~$t$, respectively. The encoder inputs the video frame~${\bf{I}}_t$ and gives the query map~${\bf{q}}_t$ of size~$H \times W \times C$, where $H$, $W$, $C$ are height, width, and the number of channels, respectively. We denote by $\mathbf{q}_t^k \in \mathbb{R}^{C}$ ($k=1,\dots K$), where $K=H\times W$, individual queries of size~$1 \times 1 \times C$ in the query map~${\bf{q}}_t$. The queries are then inputted to the memory module to read the items in the memory or to update the items, such that they record prototypical normal patterns. The detailed descriptions of the memory module are presented in the following section. The decoder inputs the queries and retrieved memory items and reconstructs the video frame~$\hat {\bf{I}}_t$.

    	\vspace{-0.4cm}
		\subsubsection{Memory}\vspace{-0.2cm}
			The memory module contains $M$~items recording various prototypical patterns of normal data. We denote by $\mathbf{p}_m \in \mathbb{R}^C$~$(m=1,\dots,M)$ the item in the memory. The memory performs reading and updating the items~(Fig.~\ref{fig:memory}). 

\vspace{-0.4cm}
			\paragraph{Read.} To read the items, we compute the cosine similarity between each query~${\mathbf{q}}_t^k$ and all memory items~$\mathbf{p}_m$, resulting in a 2-dimensional correlation map of size~$M \times K$. We then apply a softmax function along a vertical direction, and obtain matching probabilities~$w_t^{k,m}$ as follows:
				\begin{equation}\label{eq:w}
\vspace{-0.15cm}
					w_t^{k,m} = \frac{\exp((\mathbf{p}_m)^T\mathbf{q}_t^k)}{\sum_{m'=1}^M \exp((\mathbf{p}_{m'})^T\mathbf{q}_t^k)}.
\vspace{-0.1cm}
				\end{equation}
For each query~${\mathbf{q}}_t^k$, we read the memory by a weighted average of the items~$\mathbf{p}_m$ with the corresponding weights~$w_t^{k,m}$, and obtain the feature $\hat{\mathbf{p}}_t^k \in \mathbb{R}^C$ as follows:
				\begin{equation}
\vspace{-0.15cm}
                	\hat{\mathbf{p}}^k_t = \sum_{m'=1}^M w_t^{k,m'}\mathbf{p}_{m'}.
\vspace{-0.1cm}
            	\end{equation}
Using all items instead of the closest item allows our model to understand diverse normal patterns, taking into account the overall normal characteristics. That is, we represent the query~${\mathbf{q}}_t^k$ with a combination of the items~$\mathbf{p}_m$ in the memory. We apply the reading operator to individual queries, and obtain a transformed feature map~$\hat{\mathbf{p}}_t \in \mathbb{R}^{H \times W \times C}$~(\ie,~aggregated items). We concatenate it with the query map~$\mathbf{q}_t$ along the channel dimension, and input them to the decoder. This enables the decoder to reconstruct the input frame using normal patterns in the items, lessening the representation capacity of CNNs, while understanding the normality.

\vspace{-0.4cm}
		\paragraph{Update.}
For each memory item, we select all queries declared that the item is the nearest one, using the matching probabilities in~\eqref{eq:w}. Note that multiple queries can be assigned to a single item in the memory. See, for example, Fig.~\ref{fig:Confusion} in~Sec.~\ref{sec:discussion}. We denote by $U_t^m$ the set of indices for the corresponding queries for the $m$-th item in the memory. We update the item using the queries indexed by the set~$U_t^m$ only as follows:

\begin{equation}\label{eq:update}
\vspace{-0.15cm}
\mathbf{p}^m \leftarrow f(\mathbf{p}^m + \sum_{k\in U_t^m} {v'}_t^{k,m} \mathbf{q}_t^k),
\vspace{-0.1cm}
\end{equation}
 where $f(\cdot)$ is the L2 norm. By using a weighted average of the queries, rather than summing them up, we can concentrate more on the queries near the item. To this end, we compute matching probabilities~$v_t^{k, m}$ similar to~\eqref{eq:w} but by applying the softmax function to the correlation map of size $M \times K$ along a horizontal direction as 
			\begin{equation}\label{eq:v}
\vspace{-0.15cm}
				v_t^{k,m} = \frac{\exp((\mathbf{p}_m)^T\mathbf{q}_t^k)}{\sum_{k'=1}^K \exp((\mathbf{p}_m)^T\mathbf{q}_t^{k'})},
			\end{equation}
and renormalize it to consider the queries indexed by the set~$U_t^m$ as follows:
			\begin{equation}
\vspace{-0.15cm}
				{v'}_t^{k,m}=\frac{v_t^{k, m}}{\max_{k'\in U_t^m} v_t^{k', m}}.
\vspace{-0.1cm}
			\end{equation}


We update memory items recording prototypical features at both training and test time, since normal patterns in training and test sets may be different and they could vary with various factors,~\eg,~illumination and occlusion. As both normal and abnormal frames are available at test time, we propose to use a weighted regular score to prevent the memory items from recording patterns in the abnormal frames. Given a video frame~${\mathbf{I}}_t$, we use the weighted reconstruction error between~${\mathbf{I}}_t$ and $\hat{\mathbf{I}}_t$ as the regular score~$\mathcal{E}_t$: 
			\begin{equation}\label{eq:pseudo_score}
\vspace{-0.1cm}
				\mathcal{E}_t=  \sum_{i, j} W_{ij}(\hat{\mathbf{I}}_t,  \mathbf{I}_t) \|\hat{\mathbf{I}}^{ij}_t-\mathbf{I}^{ij}_t\|_2,
\vspace{-0.1cm}
			\end{equation}
	where the weight function $W_{ij}(\cdot)$ is
			\begin{equation}\label{eq:weight_function}
\vspace{-0.1cm}
				W_{ij}(\hat{\textbf{I}}_t, \textbf{I}_t) = {1-\exp(-{{||\hat{\textbf{I}}^{ij}_t-\textbf{I}^{ij}_t||}_2})\over{\sum_{i, j} 1-\exp(-{{||\hat{\textbf{I}}^{ij}_t-\textbf{I}^{ij}_t||}_2)}}},
\vspace{-0.1cm}
			\end{equation}
and $i$ and $j$ are spatial indices. When the score~$\mathcal{E}_t$ is higher than a threshold $\gamma$, we regard the frame~$\mathbf{I}_t$ as an abnormal sample, and do not use it for updating memory items. Note that we use this score only when updating the memory. The weight function allows to focus more on the regions of large reconstruction errors, as abnormal activities typically appear within small parts of the video frame.

		\vspace{-0.1cm}
	\subsection{Training loss} \label{sec:loss}
\vspace{-0.1cm}
		We exploit the video frames as a supervisory signal to discriminate normal and abnormal samples. To train our model, we use reconstruction, feature compactness, and feature separateness losses~($\mathcal{L}_\mathrm{rec}$, $\mathcal{L}_\mathrm{compact}$ and $\mathcal{L}_\mathrm{separate}$, respectively), balanced by the parameters~$\lambda_\mathrm{c}$ and $\lambda_\mathrm{s}$ as follows:
		\begin{equation}
\vspace{-0.15cm}
				\mathcal{L} = \mathcal{L}_\mathrm{rec} + \lambda_\mathrm{c}\mathcal{L}_\mathrm{compact} + \lambda_\mathrm{s}\mathcal{L}_\mathrm{separate}.
\vspace{-0.1cm}
			\end{equation}
		
\vspace{-0.3cm}
		\paragraph{Reconstruction loss.}
			The reconstruction loss makes the video frame reconstructed from the decoder similar to its ground truth by penalizing the intensity differences. Specifically, we minimize the L2 distance between the decoder output~$\hat{\bf{I}}_t$ and the ground truth~${\bf{I}}_t$:
			\begin{equation}
\vspace{-0.15cm}
				\mathcal{L}_\mathrm{rec} = \sum_{t}^{T} \|\hat{\bf{I}}_t-{\bf{I}}_t\|_2,
\vspace{-0.1cm}
			\end{equation}
			where we denote $T$ by the total length of a video sequence. We set the first time step to 1 and 5 for reconstruction and prediction tasks, respectively.
			
\vspace{-0.4cm}		
		\paragraph{Feature compactness loss.}
The feature compactness loss encourages the queries to be close to the nearest item in the memory, reducing intra-class variations. It penalizes the discrepancies between them in terms of the L2 norm as:
			\begin{equation}\label{eq:compact}
\vspace{-0.15cm}
				\mathcal{L}_\mathrm{compact}=\sum_{t}^{T} \sum_{k}^{K} \|\mathbf{q}_t^k-\mathbf{p}_{p}\|_2,
\vspace{-0.1cm}
			\end{equation}
			where $p$ is an index of the nearest item for the query $\mathbf{q}_t^k$ defined as, 
			\begin{equation}
\vspace{-0.15cm}
				p = \argmax_{m\in M} w_t^{k,m}.
			\end{equation}
Note that the feature compactness loss and the center loss~\cite{wen2016discriminative} are similar, as the memory item~$\mathbf{p}_{p}$ corresponds the center of deep features in the center loss. They are different in that the item in~\eqref{eq:compact} is retrieved from the memory, and it is updated without any supervisory signals, while the cluster center in the center loss is computed directly using the features learned from ground-truth class labels. Note also that our method can be considered as an unsupervised learning of joint clustering and feature representations. In this task, degenerate solutions are likely to be obtained~\cite{wen2016discriminative,yang2016joint}. As will be seen in our experiments, training our model using the feature compactness loss only makes all items similar, and thus all queries are mapped closely in the embedding space, losing the capability of recording diverse normal patterns. 

\vspace{-0.4cm}			
		\paragraph{Feature separateness loss.}
			Similar queries should be allocated to the same item in order to reduce the number of items and the memory size. The feature compactness loss in~\eqref{eq:compact} makes all queries and memory items close to each other, as we extract the features~(\ie,~queries) and update the items alternatively, resulting that all items are similar. The items in the memory, however, should be far enough apart from each other to consider various patterns of normal data. To prevent this problem while obtaining compact feature representations, we propose a feature separateness loss, defined with a margin of~$\alpha$ as follows:
			\begin{equation}\label{eq:separate}
\vspace{-0.15cm}
				\mathcal{L}_\mathrm{separate}= \sum_{t}^{T} \sum_k^{K} [\|\mathbf{q}_t^k - \mathbf{p}_p\|_2-\|\mathbf{q}_t^k - \mathbf{p}_n\|_2 + \alpha ]_+,
\vspace{-0.1cm}
			\end{equation}
			where we set the query~$\mathbf{q}_t^k$, its nearest item~$\mathbf{p}_p$ and the second nearest item~$\mathbf{p}_n$ as an anchor, and positive and hard negative samples, respectively. We denote by $n$ an index of the second nearest item for the query~$\mathbf{q}_t^k$:
			\begin{equation}
\vspace{-0.1cm}
				n = \argmax_{\substack{m\in M, m\neq p}} w_t^{k,m}.	
\vspace{-0.1cm}
			\end{equation}
			Note that this is different from the typical use of the triplet loss that requires ground-truth positive and negative samples for the anchor. Our loss encourages the query and the second nearest item to be distant, while the query and the nearest one to be nearby. This has the effect of placing the items far away. As a result, the feature separateness loss allows to update the item nearest to the query, whereas discarding the influence of the second nearest item, separating all items in the memory and enhancing the discriminative power.

\vspace{-0.1cm}
	\subsection{Abnormality score}\label{sec:score}
\vspace{-0.1cm}
		We quantify the extent of normalities or abnormalities in a video frame at test time. We assume that the queries obtained from a normal video frame are similar to the memory items, as they record prototypical patterns of normal data. We compute the L2 distance between each query and the nearest item as follows:
		\begin{equation}\label{eq:as_dist}
\vspace{-0.15cm}
			D(\mathbf{q}_t, \mathbf{p}) = \frac{1}{K} \sum_k^K \|\mathbf{q}_t^k-\mathbf{p}_p\|_2.
\vspace{-0.1cm}
		\end{equation}

		We also exploit the memory items implicitly to compute the abnormality score. We measure how well the video frame is reconstructed using the memory items. This assumes that anomalous patterns in the video frame are not reconstructed by the memory items. Following~\cite{liu2018future}, we compute the PSNR between the input video frame and its reonstruction:
		\begin{equation}\label{eq:as_psnr}
\vspace{-0.15cm}
			P(\hat{\mathbf{I}}_t, \mathbf{I}_t)=10\log_{10} \frac{\max (\hat{\mathbf{I}}_t)}{\|\hat{\mathbf{I}}_t-\mathbf{I}_t\|_2^2/N.}
\vspace{-0.1cm}
		\end{equation}
		where $N$ is the number of pixels in the video frame. When the frame~$\mathbf{I}_t$ is abnormal, we obtain a low value of PSNR and vice versa. Following~\cite{liu2018future,gong2019memorizing,luo2017revisit}, we normalize each error in~\eqref{eq:as_dist} and \eqref{eq:as_psnr} in the range of~[0, 1] by a min-max normalization~\cite{liu2018future}. We define the final abnormality score~$\mathcal{S}_t$ for each video frame as the sum of two metrics, balanced by the parameter~$\lambda$, as follows:
		\begin{equation}\label{eq:abnomal_score}
\vspace{-0.15cm}
			\mathcal{S}_t = \lambda (1-g(P(\hat{\mathbf{I}}_t, \mathbf{I}_t))) + (1-\lambda) g(D(\mathbf{q}_t, \mathbf{p})),
\vspace{-0.1cm}
		\end{equation}
		where we denote by $g(\cdot)$ the min-max normalization~\cite{liu2018future} over whole video frames,~\eg,
			\begin{equation}
\vspace{-0.15cm}
				g(D(\mathbf{q}_t, \mathbf{p}))=\frac{D(\mathbf{q}_t, \mathbf{p}) - \min_t (D(\mathbf{q}_t, \mathbf{p})}{\max_t (D(\mathbf{q}_t, \mathbf{p})) - \min_t (D(\mathbf{q}_t, \mathbf{p}))}.
\vspace{-0.1cm}
			\end{equation}

\vspace{-0.2cm}        
\section{Experiments}
\vspace{-0.1cm}    
	\subsection{Implementation details}
\vspace{-0.1cm}    
		\paragraph{Dataset.}
			We evaluate our method on three benchmark datasets and compare the performance with the state of the art. 1) The UCSD Ped2 dataset~\cite{li2013anomaly} contains 16 training and 12 test videos with 12 irregular events, including riding a bike and driving a vehicle. 2) The CUHK Avenue dataset~\cite{lu2013abnormal} consists of 16 training and 21 test videos with 47 abnormal events such as running and throwing stuff. 3) The ShanghaiTech dataset~\cite{luo2017revisit} contains 330 training and 107 test videos of 13 scenes. It is the largest dataset among existing benchmarks for anomaly detection. 
		
\vspace{-0.5cm}    
		\paragraph{Training.}
			We resize each video frame to the size of 256 $\times$ 256 and normalize it to the range of [-1, 1]. We set the height $H$ and the width $W$ of the query feature map, and the numbers of feature channels~$C$ and memory items~$M$ to 32, 32, 512 and 10, respectively. We use the Adam optimizer~\cite{kingma2014adam} with $\beta_1=0.9$ and $\beta_2=0.999$, with a batch size of 4 for 60, 60, and 10 epochs on UCSD Ped2~\cite{li2013anomaly}, CUHK Avenue~\cite{lu2013abnormal}, and ShanghaiTech~\cite{luo2017revisit}, respectively. We set initial learning rates to 2e-5 and 2e-4, respectively, for reconstruction and prediction tasks, and decay them using a cosine annealing method~\cite{loshchilov2016sgdr}. For the reconstruction task, we use a grid search to set hyper-parameters on the test split of UCSD Ped1~\cite{li2013anomaly}: $\lambda_\mathrm{c}= 0.01$, $\lambda_\mathrm{s}= 0.01$, $\lambda=0.7$, $\alpha=1$ and $\gamma=0.015$. We use different parameters for the prediction task similarly chosen using a grid search: $\lambda_\mathrm{c}= 0.1$, $\lambda_\mathrm{s} = 0.1$, $\lambda=0.6$, $\alpha=1$ and $\gamma=0.01$. All models are trained end-to-end using \texttt{PyTorch}~\cite{paszke2017automatic}, taking about 1, 15 and 36 hours for UCSD Ped2, CUHK Avenue, and ShanghaiTech, respectively, with an Nvidia GTX TITAN Xp.

\setlength{\tabcolsep}{0.2em}
	\begin{table}[t]
	\captionsetup{font={small}}
	\small
	\begin{center}
	\caption{Quantitative comparison with the state of the art for anomaly detection. We measure the average AUC~(\%) on UCSD Ped2~\cite{li2013anomaly}, CUHK Avenue~\cite{lu2013abnormal}, and ShanghaiTech~\cite{luo2017revisit}. Numbers in bold indicate the best performance and underscored ones are the second best.}
		\vspace{-0.3cm}
		\begin{tabular}{c | l | c | c | c}
			\hline
			\multicolumn{2}{c |}{Methods} & Ped2~\cite{li2013anomaly} & Avenue~\cite{lu2013abnormal} & Shanghai~\cite{luo2017revisit} \\
			\hline
			\multirow{7}{*}{\rb{--}} 
			& MPPCA~\cite{kim2009observe} & 69.3 & - & - \\
			& MPPC+SFA~\cite{kim2009observe} & 61.3 & - & - \\
			& MDT~\cite{mahadevan2010anomaly} & 82.9 & - & - \\
			& AMDN~\cite{xu2017detecting} & 90.8 & - & - \\
			& Unmasking~\cite{tudor2017unmasking} & 82.2 & 80.6 & - \\
			& MT-FRCN~\cite{hinami2017joint} & 92.2 & - & - \\
			& AMC~\cite{nguyen2019anomaly} & \underline{96.2} & \underline{86.9} & -\\
			\hline
			\multirow{8}{*}{\rb{Recon.~}} 
			& ConvAE~\cite{hasan2016learning} & 85.0 & 80.0 & 60.9 \\
			& TSC~\cite{luo2017revisit} & 91.0 & 80.6 & 67.9 \\
			& StackRNN~\cite{luo2017revisit} & 92.2 & 81.7 & 68.0 \\
			& AbnormalGAN~\cite{ravanbakhsh2017abnormal} & 93.5 & - & - \\
			& MemAE w/o Mem.~\cite{gong2019memorizing} & 91.7 & 81.0 & 69.7 \\
			& MemAE w/ Mem.~\cite{gong2019memorizing} & 94.1 & 83.3 & \underline{71.2} \\
			& Ours-R w/o Mem. & 86.4 & 80.6 & 65.8 \\
			& Ours-R w/ Mem. & 90.2 & 82.8 & 69.8 \\ 
			\hline
			\multirow{3}{*}{\rb{Pred.~}}
			& Frame-Pred~\cite{liu2018future}  & 95.4 & 85.1 & \textbf{72.8} \\
			& Ours-P w/o Mem. & 94.3 & 84.5 & 66.8 \\
			& Ours-P w/ Mem. & \textbf{97.0}  &  \textbf{88.5}&  70.5\\
			\hline
		\end{tabular}
		\label{table:Comparison}
	\end{center}	
		\vspace{-0.5cm}
\end{table}

	\subsection{Results}
\vspace{-0.1cm}    
		\paragraph{Comparison with the state of the art.}
			We compare in Table~\ref{table:Comparison} our models with the state of the art for anomaly detection on UCSD Ped2~\cite{li2013anomaly}, CUHK Avenue~\cite{lu2013abnormal}, and ShanghaiTech~\cite{luo2017revisit}. Following the experimental protocol in~\cite{liu2018future,gong2019memorizing,luo2017revisit}, we measure the average area under curve~(AUC) by computing the area under the receiver operation characteristics~(ROC) with varying threshold values for abnormality scores. We report the AUC performance of our models using memory modules for the tasks of frame reconstruction and future frame prediction. For comparison, we also provide the performance  without the memory module. The suffices~`-R' and `-P' indicate the reconstruction and prediction tasks, respectively. 
			
			From the table, we observe three things: (1) Our model with the prediction task~(Ours-P w/ Mem.) gives the best results on UCSD Ped2 and CUHK Avenue, achieving the average AUC of 97.0\% and 88.5\%, respectively. This demonstrates the effectiveness of our approach to exploiting a memory module for anomaly detection. Although our method is outperformed by Frame-Pred~\cite{liu2018future} on ShanghaiTech, it uses additional modules for estimating optical flow, which requires more network parameters and ground-truth flow fields. Moreover, Frame-Pred leverages an adversarial learning framework, taking lots of effort to train the network. On the contrary, our model uses a simple AE for extracting features and predicting the future frame, and thus it is much faster than Frame-Pred~(67 fps vs. 25 fps). This suggests that our model offers a good compromise in terms of AUC and runtime; (2) Our model with the reconstruction task~(Ours-R w/ Mem.) shows the competitive performance compared to other reconstructive methods on UCSD Ped2, and outperforms them on other datasets, except MemAE~\cite{gong2019memorizing}. Note that MemAE exploits 3D convolutions with 2,000 memory items of size 256. On the contrary, our model uses 2D convolutions and it requires 10 items of size 512. It is thus computationally much cheaper than MemAE:~67 fps for our model vs. 45 fps for MemAE; (3) Our memory module boosts the AUC performance significantly regardless of the tasks on all datasets. For example, the AUC gains are 2.7\%, 4.0\%, and 3.7\% on UCSD Ped2, CUHK Avenue, and ShanghaiTech, respectively, for the prediction task. This indicates that the memory module is generic and it can be added to other anomaly detection methods.


		\begin{figure}
	\captionsetup{font={small}}
  			\centering
  			\renewcommand*{\thesubfigure}{}
  			\subfigure{
    			\begin{minipage}[t]{0.30\linewidth}
					\includegraphics[width=\linewidth]{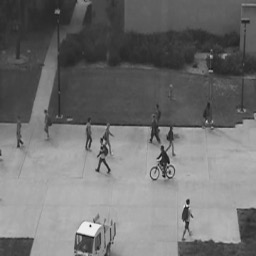}
					\includegraphics[width=\linewidth]{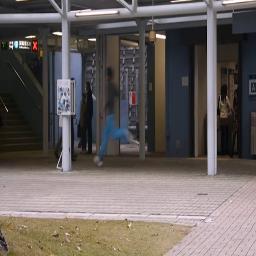}
					\includegraphics[width=\linewidth]{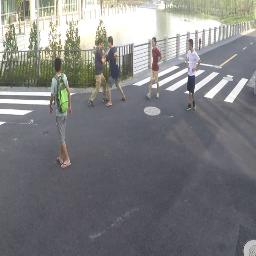}
				\end{minipage}
				}\hspace{-0.22cm} 
			\subfigure{
    			\begin{minipage}[t]{0.30\linewidth}
					\includegraphics[width=\linewidth]{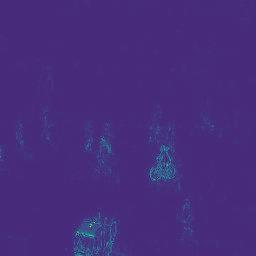}
					\includegraphics[width=\linewidth]{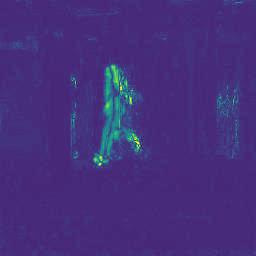}
					\includegraphics[width=\linewidth]{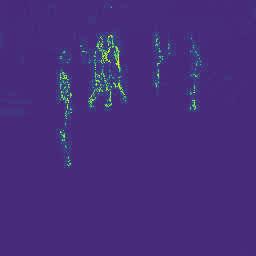}
				\end{minipage}
				}\hspace{-0.22cm}
			\subfigure{
    			\begin{minipage}[t]{0.30\linewidth}
					\includegraphics[width=\linewidth]{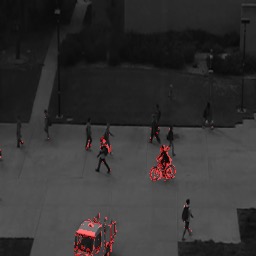}
					\includegraphics[width=\linewidth]{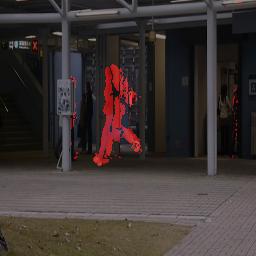}
					\includegraphics[width=\linewidth]{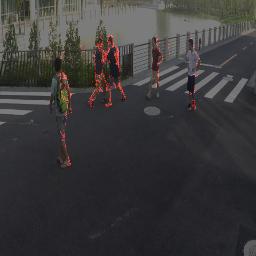}
				\end{minipage}
				}
\vspace{-0.3cm}
			\caption{Qualitative results for future frame prediction on (top to bottom) UCSD Ped2~\cite{li2013anomaly}, CUHK Avenue~\cite{lu2013abnormal}, and ShanghaiTech~\cite{luo2017revisit}: input frames~(left); prediction error~(middle); abnormal regions~(right). We can see that our model localizes the regions of abnormal events. Best viewed in color.}
			\vspace{-0.4cm}
			\label{fig:Qualitative}
		\end{figure}

\vspace{-0.4cm}			
		\paragraph{Runtime.}
			With an Nvidia GTX TITAN Xp, our current implementation takes on average 0.015 seconds to determine abnormality for an image of size 256 $\times$ 256 on UCSD Ped2~\cite{li2013anomaly}. Namely, we achieve 67 fps for anomaly detection, which is much faster than other state-of-the-art methods based on CNNs,~\eg,~20 fps for Unmasking~\cite{tudor2017unmasking}, 50 fps for StackRNN~\cite{luo2017revisit}, 25 fps for Frame-Pred~\cite{liu2018future}, and 45 fps for MemAE~\cite{gong2019memorizing} with the same setting as ours. 

\vspace{-0.4cm}
		\paragraph{Qualitative results.}
We show in Fig.~\ref{fig:Qualitative} qualitative results of our model for future frame prediction on UCSD Ped2~\cite{li2013anomaly}, CUHK Avenue~\cite{lu2013abnormal}, and ShanghaiTech~\cite{luo2017revisit}. It shows input frames, prediction error, and abnormal regions overlaid to the frame. For visualizing the anomalies, we compute pixel-wise abnormality scores similar to~\eqref{eq:abnomal_score}. We then mark the regions whose abnormality scores are larger than the average value within the frame. We can see that 1) normal regions are predicted well, while abnormal regions are not, and 2) abnormal events, such as the appearance of vehicle, jumping and fight on UCSD Ped2, CUHK Avenue, and ShanghaiTech, respectively, are highlighted.

	\subsection{Discussions}\label{sec:discussion}
\vspace{-0.1cm}
		\paragraph{Ablation study.}
			We show an ablation analysis on different components of our models in Table~\ref{table:Ablation}. We report the AUC performance for the variants of our models for reconstruction and prediction tasks on UCSD Ped2~\cite{li2013anomaly}. As the AUC performance of both tasks shows a similar trend, we describe the results for the frame reconstruction in detail. 

We train the baseline model in the first row with the reconstruction loss, and use PSNR only to compute abnormality scores. From the second row, we can see that our model with the memory module gives better results. The third row shows that the AUC performance even drops when the feature compactness loss is additionally used, as the memory items are not discriminative. The last row demonstrates that the feature separateness loss boosts the performance drastically. It provides the AUC gain of 3.8\%, which is quite significant. The last four rows indicate that 1) feature compactness and separateness losses are complementary, 2) updating the memory item using $\mathcal{E}_t$ with normal frames only at test time largely boosts the AUC performance, and 3) our abnormality score~$\mathcal{S}_t$, using both PSNR and memory items, quantifies the extent of anomalies better than the one based on PSNR only. 

		\setlength{\tabcolsep}{0.3em}
	\begin{table}
	\captionsetup{font={small}}
	\small
		\begin{center}
			\caption{Quantitative comparison for variants of our model. We measure the average AUC~(\%) on UCSD Ped2~\cite{li2013anomaly}.}
		\vspace{-0.3cm}
			\label{table:Ablation}
			\begin{tabular}{c | c | c c | c c | c}
				\hline
				 \multirow{2}*{\parbox{2em}{\centering Task}} & \multirow{1}*{\parbox{4.5em}{\centering Memory}} & \multirow{2}*{\parbox{3.5em}{\centering $\mathcal{L}_\mathrm{compact}$}} & \multirow{2}*{\parbox{3.5em}{\centering $\mathcal{L}_\mathrm{separate}$}} & \multirow{2}*{\parbox{2em}{\centering $\mathcal{E}_t$}}  & \multirow{2}*{\parbox{2em}{\centering $\mathcal{S}_t$}} & \multirow{2}*{\parbox{4em}{\centering Ped2~\cite{li2013anomaly}}}\\
				& module & & & & & \\
				\hline
				\multirow{6}{*}{\rb{Recon.~}} 
				& \xmark & - & - & - & - & 86.4 \\
				& \cmark & \xmark & \xmark & \cmark & \cmark & 86.9 \\
				& \cmark & \cmark & \xmark & \cmark & \cmark & 86.4 \\
				& \cmark & \xmark & \cmark & \cmark & \cmark & \underline{89.3} \\
				& \cmark & \cmark & \cmark & \xmark & \cmark & 87.1 \\
				& \cmark & \cmark & \cmark & \cmark & \xmark & 89.0 \\
				& \cmark & \cmark & \cmark & \cmark & \cmark & \textbf{90.2} \\
				\hline
				\multirow{6}{*}{\rb{Pred.~}}
				& \xmark & - & - & - & - & 94.3 \\
				& \cmark & \xmark & \xmark & \cmark & \cmark & 95.0 \\
				& \cmark & \cmark & \xmark & \cmark & \cmark & 94.8 \\
				& \cmark & \xmark & \cmark & \cmark & \cmark & \underline{96.5} \\
				& \cmark & \cmark & \cmark & \xmark & \cmark & 96.0 \\
				& \cmark & \cmark & \cmark & \cmark & \xmark & 95.7 \\
				& \cmark & \cmark & \cmark & \cmark & \cmark & \textbf{97.0} \\
				\hline
			\end{tabular}
		\end{center}
		\vspace{-0.6cm}
	\end{table}

\begin{figure}
	\captionsetup{font={small}}
  			\centering
  			\renewcommand*{\thesubfigure}{}
  			\subfigure{
				\includegraphics[width=0.4\linewidth]{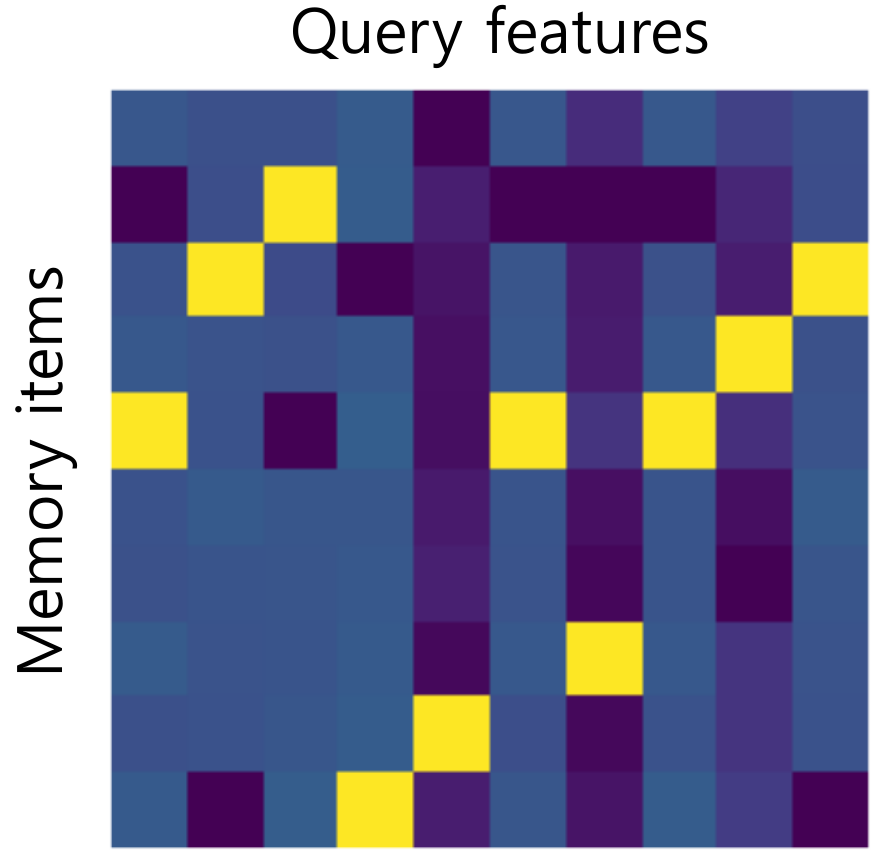}\vspace{0.25cm}
				}\hspace{-0.2cm}
			\subfigure{
				\includegraphics[width=0.4\linewidth]{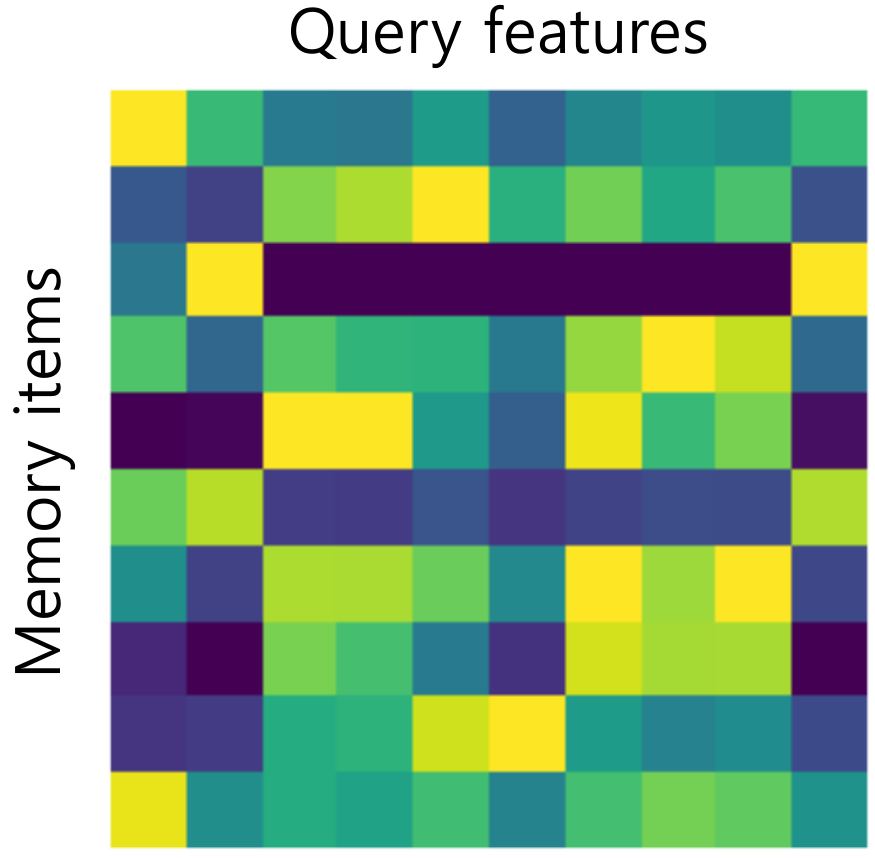}\vspace{0.25cm}
				}
		\vspace{-0.4cm}
			\caption{Visualization of matching probabilities in~\eqref{eq:w} learned with~(left) and without~(right) the feature separateness loss (blue: low, yellow: high). We randomly select 10 query features for the purpose of visualization. Best viewed in color.}
			\label{fig:Confusion}
		\vspace{-0.3cm}
		\end{figure}

\vspace{-0.5cm}		
		\paragraph{Memory items.}
We visualize in Fig.~\ref{fig:Confusion} matching probabilities in~\eqref{eq:w} from the model trained with/without the feature separateness loss for the reconstruction task on UCSD Ped2~\cite{li2013anomaly}. We observe that each query is highly activated on a few items with the separateness loss, demonstrating that the items and queries are highly discriminative, allowing the sparse access of the memory. This also indicates that abnormal samples are not likely to be reconstructed with a combination of memory items.



     \begin{figure}[t]
	\captionsetup{font={small}}
		\centering
		\renewcommand*{\thesubfigure}{}
  			\subfigure{
				\includegraphics[width=0.45\linewidth]{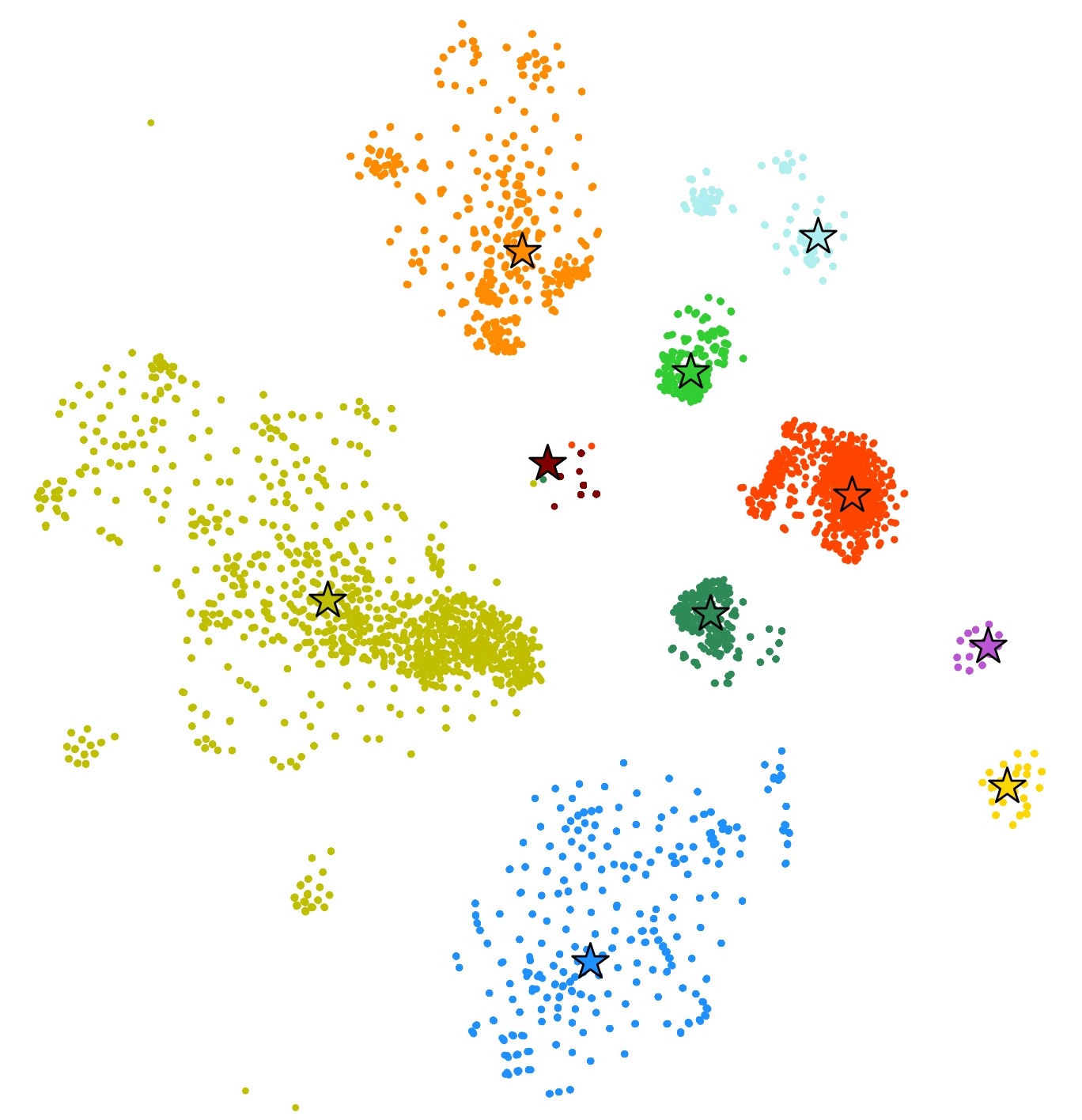}\vspace{0.25cm}
				}\hspace{-0.1cm}
			\subfigure{
				\includegraphics[width=0.45\linewidth]{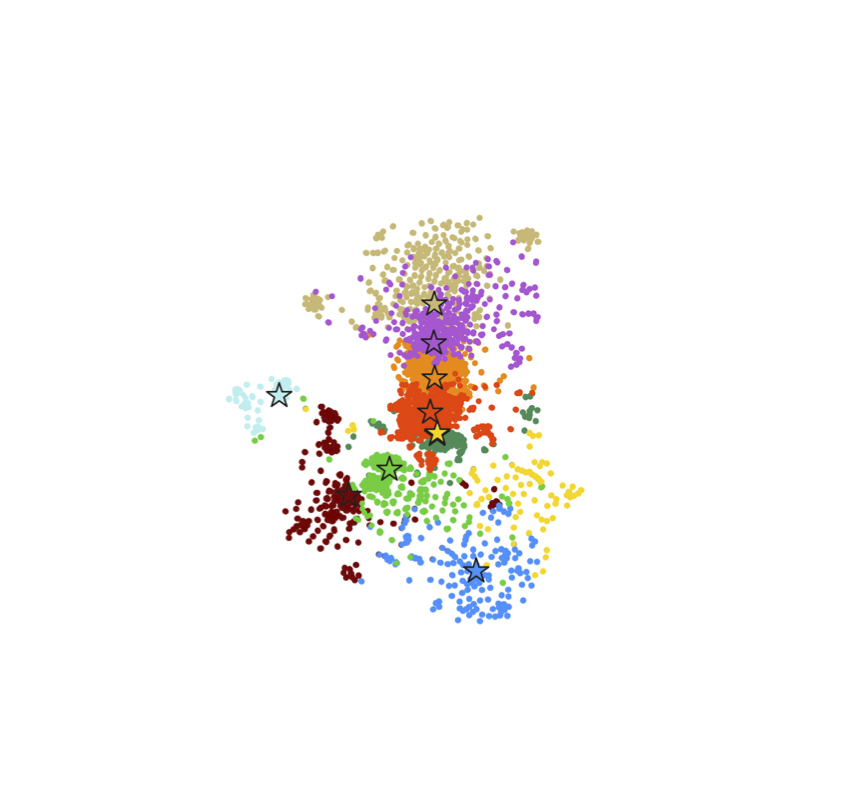}\vspace{0.25cm}
				}\hspace{-0.1cm}
		\vspace{-0.3cm}
		\caption{t-SNE~\cite{van2014accelerating} visualization for query features and memory items. We randomly sample 10K query features, learned with~(left) and without~(right) the feature separateness loss, from UCSD Ped2~\cite{li2013anomaly}. The features and memory items are shown in points and stars, respectively. The points with the same color are mapped to the same item. The feature separateness loss enables separating the items, recording the diverse prototypes of normal data. Best viewed in color.}
		\vspace{-0.4cm}
		\label{fig:tSNE}
    \end{figure}

\vspace{-0.5cm}		
		\paragraph{Feature distribution.}
		We visualize in Fig.~\ref{fig:tSNE} the distribution of query features for the reconstruction task, randomly chosen from UCSD Ped2~\cite{li2013anomaly}, learned with and without the feature separateness loss. We can see that our model trained without the separateness loss loses the discriminability of memory items, and thus all features are mapped closely in the embedding space. The separateness loss allows to separate individual items in the memory, suggesting that it enhances the discriminative power of query features and memory items significantly. We can also see that our model gives compact feature representations.

\vspace{-0.5cm}
		\paragraph{Reconstruction with motion cues.}
 Following~\cite{gong2019memorizing}, we use multiple frames for the reconstruction task. Specifically, we input sixteen successive video frames to reconstruct the ninth one. This achieves AUC of 91.0\% for UCSD Ped2, providing the AUC gain of 0.8\% but requiring more network parameters ($\sim$4MB).

			
\vspace{-0.2cm}   
\section{Conclusion}
\vspace{-0.1cm}   
	We have introduced an unsupervised learning approach to anomaly detection in video sequences that exploits multiple prototypes to consider the various patterns of normal data. To this end, we have suggested to use a memory module to record the prototypical patterns to the items in the memory. We have shown that training the memory using feature compactness and separateness losses separates the items, enabling the sparse access of the memory. We have also presented a new memory update scheme when both normal and abnormal samples exist, which boosts the performance of anomaly detection significantly. Extensive experimental evaluations on standard benchmarks demonstrate the our model outperforms the state of the art.

\vspace{-0.3cm}	
\paragraph{Acknowledgments.}This research was partly supported by Samsung Electronics Company, Ltd., Device Solutions under Grant, Deep Learning based Anomaly Detection, 2018–2020, and R\&D program for Advanced Integrated-intelligence for Identification (AIID) through the National Research Foundation of KOREA(NRF) funded by Ministry of Science and ICT (NRF-2018M3E3A1057289).

\clearpage
{\small
\bibliographystyle{ieee_fullname}
\bibliography{Anomaly_detection_camera_ready}
}

\clearpage

\end{document}